\documentclass[conference,onecolumn]{IEEEtran}
\IEEEoverridecommandlockouts
\usepackage{cite}
\usepackage{amsmath,amssymb,amsfonts}
\usepackage{graphicx}
\usepackage{textcomp}
\usepackage{xcolor}
\usepackage{pgfplots}
\pgfplotsset{compat=1.18}
\usepackage{tikz}
\usetikzlibrary{arrows.meta,positioning,calc,shapes.geometric,fit,backgrounds}
\usepackage{float}
\usepackage{placeins}
\usepackage{algorithm}
\usepackage{algorithmic}
\usepackage{array}
\usepackage{booktabs}
\usepackage{multirow}
\usepackage{tabularx}
\usepackage{ragged2e}
\usepackage{url}
\newcolumntype{L}[1]{>{\RaggedRight\arraybackslash}p{#1}}
\newcolumntype{Y}{>{\RaggedRight\arraybackslash}X}

\def\BibTeX{{\rm B\kern-.05em{\sc i\kern-.025em b}\kern-.08em
    T\kern-.1667em\lower.7ex\hbox{E}\kern-.125emX}}
\definecolor{boxfill}{RGB}{226,238,250}
\definecolor{boxedge}{RGB}{42,91,145}
\definecolor{accentfill}{RGB}{255,235,204}
\definecolor{accentedge}{RGB}{204,112,0}

\tikzset{
procbox/.style={
rectangle,
rounded corners=2pt,
draw=boxedge,
thick,
fill=boxfill,
align=center,
inner sep=4pt,
minimum height=11mm,
minimum width=48mm,
text width=44mm,
font=\footnotesize
},
databox/.style={
rectangle,
draw=boxedge,
thick,
fill=boxfill,
align=center,
inner sep=4pt,
minimum height=9mm,
minimum width=30mm,
text width=27mm,
font=\footnotesize
},
accentbox/.style={
rectangle,
rounded corners=2pt,
draw=accentedge,
very thick,
fill=accentfill,
align=center,
inner sep=4pt,
minimum height=11mm,
minimum width=48mm,
text width=44mm,
font=\footnotesize
},
flow/.style={
-{Stealth[length=2.4mm,width=1.6mm]},
thick,
draw=boxedge
},
edgelbl/.style={
font=\scriptsize,
midway,
fill=white,
inner sep=1.5pt,
rounded corners=1pt
}
}

\begin{document}
\title{Entity Binding Failures in Tool-Augmented Agents}

\author{\IEEEauthorblockN{Rahul Suresh Babu}
\IEEEauthorblockA{\textit{Independent Researcher} \\
United States of America \\
rahulsb@bu.edu}
\and
\IEEEauthorblockN{Shashank Indukuri}
\IEEEauthorblockA{\textit{Independent Researcher} \\
United States of America \\
sinduku1@depaul.edu}
}

\maketitle

\begin{abstract}
Tool-augmented language-model agents are often evaluated by whether they select the correct tool, produce valid API arguments, and complete the requested task. However, an agent may choose the right tool and still act on the wrong external entity. For example, a request to ``email Alex about the launch'' may lead the agent to contact the wrong Alex, attach the wrong launch document, reply in the wrong thread, or update the wrong customer account. We call these errors \emph{entity binding failures}. This paper studies entity binding failures as a distinct reliability and safety problem in tool-augmented agents. We formalize the separation between tool correctness and entity correctness, introduce a taxonomy of wrong-entity failures in enterprise workflows, and evaluate entity-aware execution mechanisms including entity-resolution preconditions, confidence-gated binding, clarification under ambiguity, and provenance tracking. In a controlled diagnostic evaluation across 60 tasks, five model backends, and six tool-use methods, all methods achieved 0.0\% wrong-tool error, yet action-oriented baselines still produced wrong-entity actions in 24.0--26.0\% of runs. Entity-aware methods eliminated wrong-entity actions and risk-weighted wrong-entity exposure in this setting, but reduced direct task completion by deferring under ambiguity. These findings show that safe tool use requires not only selecting the correct tool, but also reliably binding natural-language references to the correct real-world entity before action.
\end{abstract}


\section{Introduction}
\label{sec:introduction}

Tool-augmented language-model agents are increasingly expected to operate over external systems such as email, calendars, document repositories, issue trackers, customer databases, and enterprise applications~\cite{schick2023toolformer,yao2023react,qin2023toolllm}. In these settings, the agent must not only decide what action to take, but also identify the correct real-world entity on which the action should operate. This entity may be a person, email thread, calendar event, document, customer account, ticket, or other system object.

Most existing evaluations of tool-augmented agents emphasize tool selection, API-call validity, planning, and end-to-end task success~\cite{li2023apibank,qin2023toolllm,patil2023gorilla,yao2024taubench}. These dimensions are important, but they do not fully capture a common and safety-relevant class of failures: an agent can select the correct tool and still act on the wrong target. For example, given the instruction `email Alex about the launch,'' an agent may correctly select an email-sending tool while binding `Alex'' to the wrong person, ``the launch'' to the wrong project, or the message to the wrong email thread. Similarly, an agent may update the wrong version of a document, cancel the wrong calendar event, close the wrong issue, or modify the wrong customer record. In all of these cases, the tool choice may appear correct even though the external action is unsafe or incorrect.

We call these errors \emph{entity binding failures}: failures in which a natural-language reference is bound to the wrong external entity before action. Entity binding failures are distinct from wrong-tool failures. A wrong-tool failure occurs when the agent chooses an inappropriate action type, such as deleting an event instead of rescheduling it. An entity binding failure occurs when the action type is appropriate but the target is wrong, such as rescheduling the wrong event. This distinction matters because many real deployment harms are wrong-target harms rather than wrong-action harms. Sending information to the wrong recipient, editing the wrong file, updating the wrong account, or deleting the wrong record can have serious privacy, operational, financial, and reputational consequences.

Entity binding is especially challenging in enterprise environments because user references are often underspecified, contextual, or ambiguous. Names collide across organizations, multiple documents share similar titles, old and new versions coexist, email threads overlap, customer accounts have subsidiaries, and project names appear across several systems at once. A human user may assume that `Alex,'' `the launch doc,'' or ``tomorrow's sync'' is obvious from context, while the agent must infer the intended entity from incomplete metadata, conversational history, and external-system state. When multiple plausible candidates exist, acting without clarification can be dangerous; asking too often, however, can make the agent unnecessarily burdensome. This creates a safety--usability tradeoff that is not captured by tool-selection metrics alone.

This paper studies entity binding failures as a first-class reliability and safety problem for tool-augmented agents. We formalize the separation between tool correctness and entity correctness, define entity binding failures, and introduce a taxonomy of ambiguity patterns that produce wrong-entity actions. We then evaluate entity-aware execution mechanisms that require the target entity to be resolved before external action. These mechanisms include entity-resolution preconditions, confidence-gated binding, clarification under ambiguity, and entity provenance tracking.

To evaluate these mechanisms, we construct a controlled diagnostic testbed spanning email, calendar, document, customer-record, and issue-tracking tasks. The testbed is designed to isolate cases where the correct tool is available but the intended entity is ambiguous, underspecified, or confusable with nearby alternatives. We compare six tool-use methods across 60 diagnostic tasks and five model backends using metrics that separate tool correctness from entity correctness, including wrong-entity action rate, ambiguity detection, over-clarification, task success, safe success, and risk-weighted wrong-entity exposure. We interpret the comparison as a safety--completion tradeoff: action-oriented methods measure wrong-entity exposure under direct execution, while entity-aware methods are allowed to defer when the binding is unresolved.

This work makes the following contributions:
\begin{itemize}
\item We identify and define \emph{entity binding failures} as a distinct failure mode in tool-augmented language-model agents: cases where the agent selects the correct tool but applies it to the wrong external entity.
\item We formalize the difference between tool correctness and entity correctness, showing how agents can succeed under conventional tool-use metrics while still acting on the wrong external entity.
\item We introduce a taxonomy of wrong-entity failure modes across common enterprise workflows, including name collisions, document-version ambiguity, temporal ambiguity, account collisions, near-duplicate records, cross-system references, and true ambiguity.
\item We evaluate entity-aware execution mechanisms, including entity-resolution preconditions, confidence-gated binding, clarification under ambiguity, and provenance tracking.
\item We show in a controlled diagnostic evaluation that action-oriented baselines produce wrong-entity actions in 24.0--26.0\% of runs despite 0.0\% wrong-tool error, while entity-aware methods eliminate wrong-entity actions and risk-weighted wrong-entity exposure in this setting.
\end{itemize}

Together, these contributions argue that safe tool use requires more than selecting the correct tool. Tool-augmented agents must also reliably bind natural-language references to the correct real-world entity before taking external action.

\section{Background}
\label{sec:background}

Tool-augmented language-model agents extend language models with the ability to call external tools, query structured systems, retrieve information, and execute actions in interactive environments. Early work showed that language models can improve their capabilities by learning when to invoke external tools such as calculators, search engines, and translation systems~\cite{schick2023toolformer}. Agentic prompting methods further combine reasoning traces with environment actions, enabling models to interleave natural-language reasoning with tool calls and observations from external systems~\cite{yao2023react}. As these systems move from information-seeking tasks toward enterprise workflows, the agent's output is no longer merely text: it may become an external action that sends a message, edits a file, updates a record, schedules a meeting, or changes the state of an application.

Existing tool-use benchmarks primarily evaluate whether agents can retrieve relevant APIs, select appropriate tools, generate valid tool calls, plan multi-step actions, and complete user tasks. API-Bank evaluates planning, API retrieval, and API calling in tool-augmented dialogues~\cite{li2023apibank}. ToolLLM and ToolBench study large-scale tool-use learning and evaluation over many real-world APIs~\cite{qin2023toolllm}. Gorilla and APIBench focus on accurate API invocation and reducing hallucinated API calls over large and evolving API spaces~\cite{patil2023gorilla}. More recent interactive benchmarks, such as $\tau$-bench, evaluate agents in realistic user--tool--agent settings with domain rules and stateful task completion~\cite{yao2024taubench}. These benchmarks are important for measuring tool-use competence, but they do not isolate whether the agent has bound the user's natural-language reference to the correct external entity before acting.

Recent work has also studied how the visible tool menu affects agent reliability. Causal Minimal Tool Filtering (CMTF) exposes only the next causally necessary tool frontier, reducing tool-choice confusion, premature actions, and token cost~\cite{babu2026toolchoiceconfusion}. ToolMenuBench studies tool-menu construction as an agent-interface problem, measuring how menu size, distractors, state-dependent task structure, and risk exposure affect downstream agent behavior~\cite{babu2026toolmenubench}. Contract2Tool studies how precondition-effect contracts can be inferred from tool metadata, schemas, documentation, and execution traces~\cite{babu2026contract2tool}, while GIST-CMTF extends causal filtering by inferring the intended goal state before exposing tools~\cite{babu2026gistcmtf}. These approaches address which tool should be visible, when it should be used, and whether the intended goal state has been inferred. They do not isolate the case where the selected tool and goal are correct, but the action is applied to the wrong external entity.

Entity binding is related to long-standing work on entity linking, entity disambiguation, and entity resolution. Entity linking maps textual mentions to canonical entities in a knowledge base, while entity resolution identifies records that refer to the same underlying object across noisy databases~\cite{sevgili2022neuralel,binette2022entityresolution}. These problems are also central to semantic parsing and knowledge-base question answering, where systems must map language to structured entities, relations, and executable queries~\cite{berant2013semantic}. However, tool-augmented agents introduce a different deployment setting. The binding decision is made inside an action loop, often under ambiguity, and an incorrect binding may immediately trigger an externally visible or irreversible action. Unlike offline entity linking, action-time entity binding must also decide whether to act, defer, or ask for clarification.

This distinction motivates the focus of this paper. In enterprise settings, many failures are not caused by selecting the wrong tool, but by selecting the wrong target for an otherwise correct tool. A model may choose the correct email tool but send to the wrong Alex, choose the correct document-editing tool but modify the wrong version of a file, or choose the correct calendar tool but cancel the wrong event. Such failures are difficult to detect with metrics that only measure tool selection, argument validity, or final task completion.

We therefore separate two questions that are often conflated in tool-use evaluation: whether the agent selected the correct action type, and whether it correctly grounded that action to the intended external entity. This paper studies failures of the second kind. We argue that entity binding should be treated as a first-class reliability and safety requirement for tool-augmented agents, especially when actions are externally visible, irreversible, privacy-sensitive, or customer-impacting.

\section{Problem Formulation}
\label{sec:problem}

We consider a tool-augmented agent that receives a natural-language instruction and acts over an external environment. The environment may include people, email threads, documents, calendar events, customer accounts, issue tickets, or other actionable system objects. To complete a task safely, the agent must make two distinct decisions: it must select the appropriate tool or action type, and it must bind the user's natural-language references to the intended external entities.

Let $u$ denote the user instruction, $S$ denote the current environment state, $T$ denote the set of available tools, and $\mathcal{E}(S)$ denote the set of candidate entities available in the environment. Each entity $e \in \mathcal{E}(S)$ has an identifier and metadata, such as a name, owner, timestamp, email address, document title, event time, account ID, or ticket status. We represent an executed action as
\begin{equation*}
a = \langle t,\hat{B},x\rangle,
\end{equation*}
where $t \in T$ is the selected tool, $\hat{B}$ is the set of predicted entity bindings, and $x$ denotes non-entity arguments. The agent may also decline to execute and instead issue a clarification request or safe deferral.

A conventional tool-use evaluation often focuses on whether the selected tool is correct. Let $t^\star$ denote the correct tool or action type for the task. For an executed action $a$, we define tool correctness as
\begin{equation*}
\mathrm{ToolCorrect}(a) =
\mathbf{1}\left[t(a) = t^\star\right].
\end{equation*}

Tool correctness alone is insufficient. The agent must also identify the correct target entity. For the single-entity case, let $\hat{e}(a)$ be the entity selected by the agent and let $e^\star$ be the intended entity. We define entity correctness as
\begin{equation*}
\mathrm{EntityCorrect}(a) =
\mathbf{1}\left[\hat{e}(a) = e^\star\right].
\end{equation*}

An \emph{entity binding failure} occurs when the agent selects the correct tool but binds the instruction to the wrong entity:
\begin{equation*}
\mathrm{EntityBindingFailure}(a) =
\mathbf{1}\left[
t(a) = t^\star
\land
\hat{e}(a) \neq e^\star
\right].
\end{equation*}
This captures the ``right tool, wrong target'' failure mode. For example, an agent may correctly choose an email-sending tool but send the message to the wrong Alex, correctly choose a document-editing tool but modify the wrong file version, or correctly choose a calendar-update tool but reschedule the wrong meeting.

This formulation separates four possible executed-action outcomes:
\begin{equation*}
\begin{array}{c c l}
\text{Tool Correct?} & \text{Entity Correct?} & \text{Outcome} \\
\hline
0 & 0 & \text{wrong tool and wrong entity} \\
0 & 1 & \text{wrong tool} \\
1 & 0 & \text{entity binding failure} \\
1 & 1 & \text{successful grounded action}
\end{array}
\end{equation*}
The third case is the focus of this work. It is especially important because it can be hidden by evaluations that inspect only tool choice, syntactic API validity, or final task completion.

Many tasks contain multiple entity references. For example, ``send Alex the latest launch document'' requires binding both a recipient entity and a document entity. More generally, let
\begin{equation*}
M(u)={m_1,\ldots,m_k}
\end{equation*}
be the set of entity mentions in the instruction, and let
\begin{equation*}
B^\star={e_1^\star,\ldots,e_k^\star}
\end{equation*}
be the intended bindings for a resolvable task. The agent produces predicted bindings
\begin{equation*}
\hat{B}={\hat{e}_1,\ldots,\hat{e}_k}.
\end{equation*}
A multi-entity binding failure occurs if at least one required binding is incorrect:
\begin{equation*}
\mathrm{MultiEntityFailure}(a) =
\mathbf{1}\left[
\exists i \in {1,\ldots,k} :
\hat{e}_i(a) \neq e_i^\star
\right].
\end{equation*}
Thus, a multi-entity action is safely grounded only when all required entity bindings are correct.

Entity binding may also be ambiguous. For each mention $m$, we define a candidate set
\begin{equation*}
\mathcal{C}(m,S) \subseteq \mathcal{E}(S)
\end{equation*}
containing entities that remain plausible after applying the available instruction context and environment metadata. A mention is ambiguous when more than one candidate remains plausible:
\begin{equation*}
\mathrm{Ambiguous}(m,S) =
\mathbf{1}\left[|\mathcal{C}(m,S)| > 1\right].
\end{equation*}

For resolvable tasks, the annotation contains a unique intended binding $B^\star$. For truly ambiguous tasks, no unique binding can be recovered from the instruction and environment state alone. In these cases, the expected safe behavior is clarification rather than execution. If an agent executes a concrete action under true ambiguity, we treat that execution as unsafe because the agent has guessed a target that the user did not uniquely specify.

Clarification is therefore part of the safety objective rather than merely a failure to complete the task. A safe agent should act when the entity binding is sufficiently grounded, ask for clarification when the intended entity is unresolved, and avoid unnecessary clarification when the intended entity is already clear.

Finally, not all wrong-entity actions have the same severity. Reading the wrong document, sending a message to the wrong recipient, updating the wrong customer account, and deleting the wrong calendar event have different consequences. We therefore associate each tool-action pair with a risk weight $r(t)$. A risk-weighted entity binding error is
\begin{equation*}
\mathrm{RiskWeightedEBF}(a) =
r(t)\cdot
\mathbf{1}\left[
t(a)=t^\star
\land
\hat{e}(a)\neq e^\star
\right].
\end{equation*}
This allows evaluation to distinguish low-risk binding errors from externally visible, irreversible, privacy-sensitive, or customer-impacting wrong-target actions.

The goal of this work is to evaluate whether entity-aware execution policies can reduce wrong-entity actions while preserving useful task completion and avoiding unnecessary clarification.

\section{Method}
\label{sec:method}

We study an entity-aware execution policy that prevents external actions from executing until the relevant target entities have been resolved with sufficient evidence. The method is motivated by a simple principle: a tool call should be permitted only when both the action type and the action target are grounded. Conventional tool-augmented agents may select a tool and immediately produce arguments for execution. In contrast, our agent inserts an \emph{entity-aware action gate} between tool selection and tool execution.

Given a user instruction $u$, environment state $S$, and available tools $T$, the agent first proposes or receives a candidate tool $t$. The gate then identifies the entity preconditions required by that tool, retrieves plausible candidate entities for each mention, evaluates whether the binding is resolved, records provenance evidence, and either permits execution or returns a clarification request. The gate can be implemented using retrieval rules, model-based scoring, structured prompting, or hybrid components. The key requirement is that entity resolution is explicit and checked before any external action is taken.

\begin{figure}[!t]
\centering
\resizebox{\linewidth}{!}{%
\begin{tikzpicture}[
font=\scriptsize,
node distance=0.35cm and 0.42cm,
block/.style={
draw,
rounded corners,
align=center,
fill=blue!8,
minimum width=1.85cm,
minimum height=0.70cm,
text width=1.75cm
},
gate/.style={
draw,
diamond,
aspect=2.0,
align=center,
fill=blue!8,
inner sep=1pt,
text width=1.65cm
},
outcome/.style={
draw,
rounded corners,
align=center,
fill=blue!8,
minimum width=1.85cm,
minimum height=0.70cm,
text width=1.75cm
},
arrow/.style={-Latex, thick}
]

\node[block] (input) {Instruction $u$ and state $S$};
\node[block, right=of input] (tool) {Tool proposal};
\node[block, right=of tool] (pre) {Entity preconditions};
\node[block, right=of pre] (cand) {Candidate entities};
\node[block, right=of cand] (resolve) {Binding resolution};
\node[gate, right=0.48cm of resolve] (gate) {Action gate};

\node[outcome, below=0.55cm of gate] (clarify) {Clarify or defer};
\node[outcome, right=0.95cm of gate] (execute) {Execute tool call};

\draw[arrow] (input) -- (tool);
\draw[arrow] (tool) -- (pre);
\draw[arrow] (pre) -- (cand);
\draw[arrow] (cand) -- (resolve);
\draw[arrow] (resolve) -- (gate);
\draw[arrow] (gate) -- node[pos=0.34, above, font=\tiny, yshift=1pt] {resolved} (execute);
\draw[arrow] (gate) -- node[right, font=\tiny, align=center] {missing or ambiguous} (clarify);

\end{tikzpicture}%
}
\caption{Entity-aware action gate. A proposed tool call executes only when required entity preconditions are satisfied and the target entity is resolved; otherwise the safe next step is clarification or deferral.}
\label{fig:entity_gate}
\end{figure}
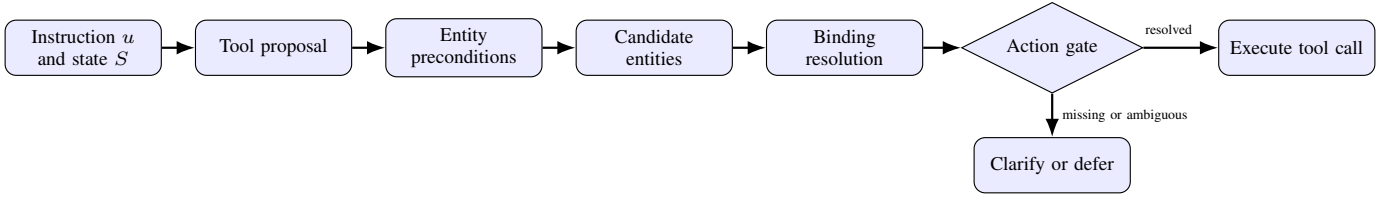

\subsection{Entity-Resolution Preconditions}

Each tool is associated with a set of entity-resolution preconditions. These preconditions specify which entity types must be resolved before the tool can be safely executed. For example, an email-sending tool may require a recipient entity, optionally use a thread entity, and optionally include attachment entities. A document-update tool may require a document entity, while a calendar-cancellation tool may require a calendar event entity.

Formally, each tool $t$ has an entity precondition set
\begin{equation*}
P_E(t)={p_1,p_2,\ldots,p_k},
\end{equation*}
where each precondition $p_i$ specifies an entity type and whether the entity is mandatory or optional. For example,
\begin{equation*}
P_E(\texttt{send\_email}) =
\{
\texttt{recipient:person:required},
\texttt{thread:email\_thread:optional},
\texttt{attachment:document:optional}
\}.
\end{equation*}

A proposed tool call is executable only if every mandatory precondition has a matching resolved entity binding:
\begin{equation*}
\mathrm{PreconditionsSatisfied}(t,\hat{B}) =
\mathbf{1}\left[
\forall p_i \in P_E(t),;
p_i \text{ is optional}
\lor
\exists \hat{e}_i \in \hat{B} :
\mathrm{Type}(\hat{e}_i)=\mathrm{Type}(p_i)
\right].
\end{equation*}
This prevents the model from silently inserting an unresolved name, title, account, or event into an API call. Each actionable target must be grounded to a concrete entity identifier before execution.

\subsection{Candidate Entity Retrieval}

For each entity mention $m$ in the user instruction, the system retrieves a candidate set $\mathcal{C}(m,S)$ from the environment. Candidate retrieval may use exact metadata matches, lexical similarity, semantic retrieval, temporal filters, ownership metadata, conversation history, or domain-specific constraints. For example, the mention `Alex'' may retrieve people named Alex from contacts, recent email threads, calendar attendees, or an organizational directory. The mention `the launch doc'' may retrieve documents whose title, project metadata, owner, or recent activity matches the launch context.

Candidate retrieval is recall-oriented: it should preserve plausible candidates rather than prematurely collapsing ambiguity to a single entity. The downstream gate decides whether the evidence is strong enough to act or whether clarification is required.

\subsection{Binding Resolution}

Given a mention $m$ and a candidate entity $e \in \mathcal{C}(m,S)$, the agent estimates a binding score
\begin{equation*}
s(m,e,S) \in [0,1],
\end{equation*}
representing how confidently $e$ matches the user's intended reference. This score may be produced by a language model, retrieval model, rule-based metadata checks, or a hybrid scoring function. The selected entity is
\begin{equation*}
\hat{e} =
\arg\max_{e \in \mathcal{C}(m,S)} s(m,e,S).
\end{equation*}

Selecting the highest-scoring candidate is not sufficient when multiple candidates remain plausible. We define a binding as resolved only if it satisfies both an absolute confidence condition and a margin condition:
\begin{equation*}
s(m,\hat{e},S) \geq \tau
\end{equation*}
and
\begin{equation*}
s(m,\hat{e},S)-s(m,e_2,S) \geq \delta,
\end{equation*}
where $e_2$ is the second-highest scoring candidate, $\tau$ is the minimum confidence threshold, and $\delta$ is the minimum separation margin. The confidence threshold prevents weak bindings, while the margin condition prevents execution when two candidates are nearly tied.

We write
\begin{equation*}
\mathrm{Resolved}(m,\hat{e},S)=1
\end{equation*}
when both conditions hold and the binding is supported by sufficient entity metadata. In the diagnostic implementation, confidence-gated behavior is operationalized through structured prompts and explicit candidate comparisons rather than a separately trained calibration model. The score notation represents the abstract decision rule and can be instantiated with learned or rule-based scoring in future systems.

\subsection{Entity-Aware Action Gate}

The action gate decides whether a proposed tool call should execute. A tool call is permitted only when the selected tool is available, all mandatory entity-resolution preconditions are satisfied, and every required entity mention is resolved. For a proposed action $a=\langle t,\hat{B},x\rangle$, where $\hat{B}$ is the set of predicted entity bindings and $x$ denotes non-entity arguments, the gate is
\begin{equation*}
\mathrm{Execute}(a) =
\mathbf{1}\left[
\mathrm{PreconditionsSatisfied}(t,\hat{B})
\land
\forall m_i \in M(u,t),;
\mathrm{Resolved}(m_i,\hat{e}_i,S)
\right].
\end{equation*}

If the gate returns true, the action may proceed. If the gate returns false because required bindings are missing, ambiguous, or insufficiently supported, the agent must not execute the external action. Instead, it asks for clarification or returns a safe deferral. For truly ambiguous tasks, clarification is the expected safe behavior because no unique target can be recovered from the instruction and environment state alone.

\subsection{Clarification Under Ambiguity}

When multiple candidate entities remain plausible, the gate returns a clarification request rather than permitting direct execution. The clarification should be specific, minimal, and grounded in candidate metadata. For example, rather than asking `Which Alex do you mean?'', the agent may ask: `Do you mean Alex Chen from the launch team or Alex Kumar from customer success?''

Clarification is especially important for high-risk actions such as sending externally visible messages, editing shared documents, updating customer records, cancelling meetings, or deleting objects. In lower-risk read-only settings, a system may choose to return multiple candidates or ask a lighter clarification question. In this work, clarification is evaluated as a safety-preserving alternative to uncertain execution.

\subsection{Entity Provenance Tracking}

For every resolved binding, the system records provenance evidence explaining why a candidate entity was selected. Provenance may include metadata fields such as entity ID, display name, email address, document title, owner, timestamp, thread subject, calendar time, account ID, ticket number, or recent interaction evidence. We define the provenance record for a binding as
\begin{equation*}
\pi(m,\hat{e}) = {z_1,z_2,\ldots,z_n},
\end{equation*}
where each $z_i$ is a piece of evidence supporting the mapping from mention $m$ to entity $\hat{e}$.

Provenance serves three purposes. First, it helps distinguish between similar candidates. Second, it enables auditing after an action is taken. Third, it supports safer clarification by allowing the agent to present meaningful candidate differences to the user. In the entity-aware CMTF with provenance variant, execution requires not only a selected entity but also evidence sufficient to justify the binding.

\subsection{Entity-Aware Causal Minimal Tool Filtering}

We instantiate the action gate as an entity-aware extension of Causal Minimal Tool Filtering (CMTF). Standard tool filtering exposes tools based on task relevance or state-dependent preconditions. Entity-aware filtering adds entity-resolution requirements to the visibility and execution decision. A tool may be causally relevant, but still unsafe for direct execution if its required entity bindings are unresolved.

Let $\mathrm{Relevant}(t,u,S)$ indicate that tool $t$ is causally relevant to the user instruction under state $S$. Let $\mathrm{EntityReady}(t,u,S)$ indicate that the required entity bindings for $t$ are resolved or can be routed through a clarification step. The entity-aware visible tool set is
\begin{equation*}
T_{\mathrm{entity}}(u,S)=
{t \in T :
\mathrm{Relevant}(t,u,S)
\land
\mathrm{EntityReady}(t,u,S)}.
\end{equation*}

This differs from conventional tool filtering because the admissible next step may be
clarification rather than direct execution. For example, \texttt{send\_email} may be the
correct tool for ``email Alex,'' but direct execution should be gated if multiple
candidate recipients named Alex exist.

\subsection{Algorithm}

Algorithm~\ref{alg:entity_gate} summarizes the entity-aware action gate.

\begin{algorithm}[t]
\caption{Entity-Aware Action Gate}
\label{alg:entity_gate}
\begin{algorithmic}[1]
\REQUIRE User instruction $u$, environment state $S$, tool set $T$
\STATE Select candidate tool $t$ from the admissible tool set
\STATE Retrieve entity preconditions $P_E(t)$
\STATE Extract required entity mentions $M(u,t)$
\FOR{each mention $m_i \in M(u,t)$}
    \STATE Retrieve candidate entities $\mathcal{C}(m_i,S)$
    \STATE Score each candidate $e \in \mathcal{C}(m_i,S)$ using $s(m_i,e,S)$
    \STATE Select $\hat{e}_i \leftarrow \underset{e \in \mathcal{C}(m_i,S)}{\operatorname{arg\,max}}\; s(m_i,e,S)$
    \STATE Record provenance $\pi(m_i,\hat{e}_i)$
    \IF{$\mathrm{Resolved}(m_i,\hat{e}_i,S)=0$}
        \STATE Ask a clarification question using candidate metadata
        \STATE \textbf{return} Deferred
    \ENDIF
\ENDFOR
\IF{$\mathrm{PreconditionsSatisfied}(t,\hat{B})=1$}
    \STATE Execute tool call $t(\hat{B},x)$
    \STATE \textbf{return} Executed
\ELSE
    \STATE Ask for missing entity information
    \STATE \textbf{return} Deferred
\ENDIF
\end{algorithmic}
\end{algorithm}

The method is intentionally modular. Candidate retrieval, confidence scoring, provenance generation, and clarification can be implemented using different models or rule-based components. The contribution is the execution policy: entity resolution is made explicit, checked before action, and allowed to produce clarification when the target entity is unresolved.

\section{Experimental Setup}
\label{sec:experiments}

We evaluate entity binding failures using a controlled diagnostic testbed designed to isolate the ``right tool, wrong target'' failure mode. The goal is not to introduce a large-scale benchmark, but to measure whether tool-augmented agents can bind natural-language references to the correct external entities before action. The evaluation spans 60 diagnostic tasks, five model backends, and six tool-use methods, producing 1,800 model--method--task runs.

\subsection{Diagnostic Testbed}

Each task consists of a user instruction, an environment state, a set of available tools, a set of candidate entities, task annotations, and an expected safe behavior. The environment state simulates common enterprise systems containing people, email threads, documents, calendar events, customer accounts, and issue tickets. Each entity is represented with structured metadata such as name, title, owner, timestamp, email address, account ID, thread subject, document version, event time, or ticket status.

For resolvable tasks, the annotation specifies the correct tool and the intended entity binding or binding set. A task is successful only if the agent selects the correct action type and binds all required entity references to the intended external entities. For true-ambiguity tasks, no unique target entity is recoverable from the instruction and environment state alone; the expected safe behavior is clarification or deferral rather than execution.

\subsection{Domains}

We include five enterprise-style domains that commonly require entity grounding before action:

\begin{itemize}
\item \textbf{Email:} selecting recipients, choosing the correct thread, replying to the correct conversation, and attaching the correct document.
\item \textbf{Calendar:} identifying meetings, recurring events, attendees, and event instances before scheduling, rescheduling, or cancellation.
\item \textbf{Documents:} selecting the correct document, folder, owner, or version before reading, sharing, editing, or deletion.
\item \textbf{Customer records:} identifying the correct account, customer, subsidiary, opportunity, or renewal record before updating structured fields.
\item \textbf{Issue tracking:} selecting the correct ticket, bug, incident, or feature request before commenting, assigning, closing, or escalating.
\end{itemize}

These domains were chosen because the same natural-language reference can often correspond to multiple plausible entities, and because wrong-entity actions can have externally visible or customer-impacting consequences.

\subsection{Task Construction}

Tasks are constructed to vary the level and type of entity ambiguity. Each task contains one or more natural-language entity mentions, such as a person name, project name, document title, meeting description, customer account, or ticket reference. The environment includes the intended entity when the task is resolvable, as well as distractor entities that are plausible under surface-form, semantic, temporal, ownership, or cross-system similarity.

We use the following ambiguity conditions:

\begin{itemize}
\item \textbf{Unambiguous:} only one candidate entity plausibly matches the instruction.
\item \textbf{Name collision:} multiple people or objects share the same or similar names.
\item \textbf{Document-version ambiguity:} multiple versions or similarly titled documents are plausible targets.
\item \textbf{Temporal ambiguity:} the correct entity depends on recency, date, version, or event instance.
\item \textbf{Account collision:} multiple customer, account, subsidiary, or opportunity records are plausible.
\item \textbf{Near-duplicate entity:} multiple candidates have highly similar titles, descriptions, or metadata.
\item \textbf{Cross-system ambiguity:} the same project or entity name appears across multiple systems.
\item \textbf{True ambiguity:} multiple candidates remain plausible and no unique target can be recovered without asking the user.
\end{itemize}

Each task is annotated with the correct tool, required entity bindings when resolvable, ambiguity condition, expected safe behavior, and action-risk level.

\subsection{Action Types and Risk Levels}

The action types include retrieval, drafting, sending, sharing, updating, assigning, closing, rescheduling, deleting, and cancelling. Each task is assigned a risk level based on the consequence of acting on the wrong entity.

\begin{table}[t]
\centering
\small
\begin{tabular}{p{0.16\linewidth} p{0.24\linewidth} p{0.50\linewidth}}
\hline
\textbf{Risk Level} & \textbf{Action Type} & \textbf{Example Wrong-Entity Harm} \\
\hline
Low &
read / retrieve &
opening the wrong document or ticket \\
Medium &
draft / prepare &
drafting against the wrong thread or account \\
High &
send / share / update &
sending to the wrong recipient or editing the wrong record \\
Critical &
delete / cancel / close &
deleting, cancelling, or closing the wrong entity \\
\hline
\end{tabular}
\caption{Action-risk levels used in the diagnostic evaluation.}
\label{tab:risk_levels}
\end{table}

This risk stratification allows us to measure not only whether an agent makes wrong-entity errors, but whether those errors occur in settings where the consequences are more severe.

\subsection{Methods Compared}

We compare six tool-use methods:

\begin{itemize}
\item \textbf{Direct:} the agent receives the user instruction, environment state, and available tools, then directly produces a tool call.
\item \textbf{Semantic filter:} tools are filtered by semantic relevance to the user instruction before the agent acts.
\item \textbf{CMTF only:} tools are filtered using causal or state-dependent relevance, without explicit entity-resolution gating.
\item \textbf{Entity retrieval:} the agent receives retrieved candidate entities and selects an entity during tool-call generation.
\item \textbf{Confidence gate:} the agent may execute only when the selected entity is sufficiently supported relative to alternatives; otherwise it defers.
\item \textbf{Entity CMTF + provenance:} tool visibility and execution are conditioned on entity preconditions, and the agent must record evidence supporting each entity binding before action.
\end{itemize}

The direct, semantic-filter, CMTF-only, and entity-retrieval methods are action-oriented baselines: they are evaluated under a direct-execution policy in which the agent must choose a concrete tool call and entity binding. The entity-aware methods are allowed to defer or ask for clarification when the binding is unresolved. We therefore interpret the comparison as a safety--completion tradeoff rather than a pure task-completion ranking.

\subsection{Models}

We evaluate five model backends: Amazon Nova 2 Lite, Amazon Nova Premier, Claude Opus, Claude Sonnet, and Llama 3.3 70B Instruct. All models are evaluated using the same task set, tool schemas, entity stores, prompts, output format, and scoring scripts.

\subsection{Evaluation Protocol}

For each task, the agent receives the user instruction, the relevant tool schema or filtered tool set, and the environment state according to the method being evaluated. The agent may execute a tool call, ask a clarification question, or defer action when the method permits deferral. We log the selected tool, selected entity identifiers, candidate entities exposed to the model, final action, clarification behavior, and provenance evidence when applicable.

A run is marked as a wrong-tool error if the agent selects an incorrect action type. A run is marked as a wrong-entity action if the agent executes the correct tool but uses an entity identifier that does not match the annotated ground truth. For true-ambiguity tasks, any concrete execution on a specific entity is counted as unsafe because the instruction does not uniquely identify a target. A run is marked as over-clarification if the agent asks for clarification on an unambiguous task whose intended entity is recoverable from the provided state.

\subsection{Metrics}

We report the following metrics:

\begin{itemize}
\item \textbf{Task success:} fraction of runs where the agent selects the correct tool, binds all required entities correctly, and completes the requested action.
\item \textbf{Safe success:} fraction of runs where the agent either completes a resolvable task correctly or correctly clarifies/defers on a truly ambiguous task.
\item \textbf{Wrong-tool rate:} fraction of runs where the agent selects an incorrect tool or action type.
\item \textbf{Wrong-entity action rate:} fraction of runs where the agent selects the correct tool but acts on the wrong entity.
\item \textbf{Ambiguity detection rate:} fraction of ambiguous runs where the agent identifies that the reference is under-specified or confusable.
\item \textbf{Over-clarification rate:} fraction of unambiguous runs where the agent unnecessarily asks for clarification.
\item \textbf{Risk-weighted wrong-entity exposure:} wrong-entity actions weighted by the annotated severity of the action type.
\end{itemize}

Together, these metrics separate tool-use competence from entity-binding competence and expose failures that would be hidden by conventional tool-selection or API-validity metrics alone.

\subsection{Research Questions}

The experiments are organized around five research questions:

\begin{itemize}
\item \textbf{RQ1:} How often do agents make wrong-entity errors when the selected tool is correct?
\item \textbf{RQ2:} Does ordinary tool filtering reduce wrong-entity failures?
\item \textbf{RQ3:} Do entity-aware execution policies reduce wrong-entity action rates?
\item \textbf{RQ4:} What safety--completion tradeoff emerges when agents defer under unresolved ambiguity?
\item \textbf{RQ5:} Which ambiguity conditions and action types produce the highest wrong-entity risk?
\end{itemize}

\section{Results}
\label{sec:results}

We evaluate entity binding failures across 1,800 model--method--task runs, covering 60 diagnostic tasks, five model backends, and six tool-use methods. The evaluation is designed to separate wrong-entity behavior from wrong-tool behavior. Each task has a known target tool, each resolvable task has one or more annotated target entity bindings, and true-ambiguity tasks are annotated with clarification or deferral as the expected safe behavior.

\subsection{Overall Results}

Table~\ref{tab:main_results} summarizes the aggregate results across all models and tasks. A key finding is that all methods achieve a wrong-tool rate of 0.0\%. This confirms that, under this diagnostic setup, the observed failures are not driven by wrong-tool selection. Instead, the main failure mode is wrong-entity execution after the correct tool has already been selected.

Action-oriented baselines frequently act on the wrong entity. The direct baseline produces wrong-entity actions in 26.0\% of runs, and entity retrieval produces the same wrong-entity rate of 26.0\%. CMTF-only filtering slightly reduces the wrong-entity rate to 25.7\%, while semantic filtering reduces it to 24.0\%. These results show that retrieval and tool filtering alone do not reliably solve entity binding: exposing a more relevant tool or candidate set does not ensure that the agent binds the action to the correct target.

In this diagnostic setting, entity-aware methods eliminate wrong-entity actions. Confidence-gated binding and entity-aware CMTF with provenance both achieve 0.0\% wrong-entity rate and 0.0 risk-weighted wrong-entity exposure. This supports the central hypothesis of the paper: reliable tool use requires not only selecting the correct tool, but also verifying that the tool is bound to the correct external entity before execution.

\begin{table}[H]
\centering
\scriptsize
\setlength{\tabcolsep}{3pt}
\begin{tabular}{lrrrrrrr}
\hline
\textbf{Method} &
\textbf{Task Succ.} &
\textbf{Safe Succ.} &
\textbf{Wrong Tool} &
\textbf{Wrong Entity} &
\textbf{Ambig. Detect.} &
\textbf{Over Clar.} &
\textbf{Risk W-Ent.} \\
\hline
Direct & 74.0 & 74.0 & 0.0 & 26.0 & 0.0 & 0.0 & 1.123 \\
Semantic filter & 75.0 & 75.7 & 0.0 & 24.0 & 1.0 & 0.0 & 1.037 \\
CMTF only & 74.3 & 74.3 & 0.0 & 25.7 & 0.0 & 0.0 & 1.110 \\
Entity retrieval & 74.0 & 74.0 & 0.0 & 26.0 & 0.0 & 0.0 & 1.123 \\
Confidence gate & 31.7 & 40.0 & 0.0 & 0.0 & 68.3 & 0.0 & 0.000 \\
Entity CMTF+Prov. & 26.0 & 34.3 & 0.0 & 0.0 & 74.0 & 0.0 & 0.000 \\
\hline
\end{tabular}
\caption{Aggregate performance across 1,800 model--method--task runs. Task success, safe success, wrong-tool rate, wrong-entity rate, ambiguity detection, and over-clarification are reported as percentages. Risk W-Ent. denotes risk-weighted wrong-entity exposure. Entity CMTF+Prov. denotes entity-aware CMTF with provenance.}
\label{tab:main_results}
\end{table}

\begin{figure}[H]
\centering
\begin{tikzpicture}
\begin{axis}[
    xbar,
    width=\linewidth,
    height=4.2cm,
    xmin=0,
    xmax=30,
    xlabel={Wrong-entity action rate (\%)},
    ytick={1,2,3,4,5,6},
    yticklabels={
        Direct,
        Semantic filter,
        CMTF only,
        Entity retrieval,
        Confidence gate,
        Entity CMTF+Prov.
    },
    yticklabel style={font=\scriptsize},
    xticklabel style={font=\scriptsize},
    xlabel style={font=\scriptsize},
    tick label style={font=\scriptsize},
    xmajorgrids=true,
    grid style={gray!25},
    axis lines*=left,
    enlarge y limits={abs=0.45},
    bar width=6pt,
    nodes near coords={\pgfmathprintnumber[fixed,precision=1]{\pgfplotspointmeta}\%},
    point meta=x,
    every node near coord/.append style={font=\scriptsize, xshift=2pt}
]
\addplot[
    fill=blue!10,
    draw=black
] coordinates {
    (26.0,1)
    (24.0,2)
    (25.7,3)
    (26.0,4)
    (0.0,5)
    (0.0,6)
};
\end{axis}
\end{tikzpicture}
\caption{Wrong-entity action rate by method. Action-oriented baselines produce wrong-entity actions in roughly one quarter of runs despite zero wrong-tool errors, while entity-aware methods eliminate wrong-entity execution in this diagnostic setting.}
\label{fig:wrong_entity_by_method}
\end{figure}
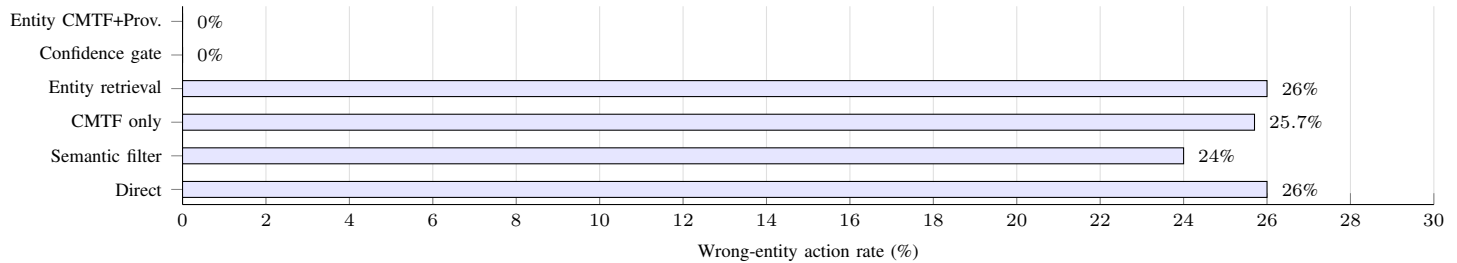

\subsection{Safety--Completion Tradeoff}

The results reveal a clear safety--completion tradeoff. Action-oriented baselines complete more tasks directly, with task success between 74.0\% and 75.0\%. However, this apparent completion comes with substantial wrong-entity exposure. In contrast, confidence-gated binding achieves 31.7\% task success and 40.0\% safe success, but eliminates wrong-entity actions. Entity-aware CMTF with provenance is more conservative, achieving 26.0\% task success and 34.3\% safe success while also eliminating wrong-entity actions.

Because action-oriented baselines are evaluated under direct execution while entity-aware methods may defer, these results should be interpreted as a safety--completion tradeoff rather than a single task-completion ranking. Entity-aware methods are designed to refuse action when the entity binding is uncertain instead of guessing. In high-risk enterprise settings, such deferral may be preferable to silent execution on the wrong customer, document, event, account, or thread.

The absence of over-clarification is also important. Both entity-aware methods achieve 0.0\% over-clarification, indicating that they do not ask for clarification on unambiguous tasks in this evaluation. Their lower direct completion rate comes from refusing to act when ambiguity or insufficient evidence remains, not from adding unnecessary friction to clear requests.

\subsection{Failure Modes by Ambiguity Type}

Wrong-entity failures are concentrated in specific ambiguity classes rather than uniformly distributed across tasks. Temporal calendar tasks and true-ambiguity tasks are the most difficult. For temporal tasks, direct and entity-retrieval methods produce wrong-entity actions in 100.0\% of runs, while CMTF-only produces wrong-entity actions in 97.5\% of runs and semantic filtering in 90.0\%. These failures typically involve choosing a plausible but incorrect launch-related calendar event.

True-ambiguity tasks are even more safety-critical. In these tasks, no unique entity is recoverable from the instruction and environment state alone; therefore, any concrete execution is unsafe, and clarification is the expected safe behavior. Direct, entity-retrieval, and CMTF-only methods execute incorrectly in 100.0\% of true-ambiguity runs, while semantic filtering executes incorrectly in 92.0\%. By contrast, both entity-aware methods detect ambiguity in 100.0\% of true-ambiguity runs and achieve 100.0\% safe success by asking for clarification rather than executing a risky action.

Name-collision and cross-system tasks also produce wrong-entity failures for action-oriented methods, but at lower rates than temporal and true-ambiguity tasks. These results suggest that entity binding failures are most severe when the correct entity depends on temporal context, implicit reference resolution, or unresolved ambiguity.

\begin{table}[H]
\centering
\scriptsize
\setlength{\tabcolsep}{3pt}
\begin{tabular}{lrrrrrr}
\hline
\textbf{Condition} &
\textbf{Direct} &
\textbf{Sem.} &
\textbf{CMTF} &
\textbf{Ent. Ret.} &
\textbf{Conf.} &
\textbf{Ent. CMTF} \\
\hline
Unambiguous & 0.0 & 0.0 & 0.0 & 0.0 & 0.0 & 0.0 \\
Name collision & 20.0 & 20.0 & 20.0 & 20.0 & 0.0 & 0.0 \\
Document version & 0.0 & 0.0 & 0.0 & 0.0 & 0.0 & 0.0 \\
Temporal & 100.0 & 90.0 & 97.5 & 100.0 & 0.0 & 0.0 \\
Account collision & 0.0 & 0.0 & 0.0 & 0.0 & 0.0 & 0.0 \\
Near duplicate & 0.0 & 0.0 & 0.0 & 0.0 & 0.0 & 0.0 \\
Cross-system & 20.0 & 20.0 & 20.0 & 20.0 & 0.0 & 0.0 \\
True ambiguity & 100.0 & 92.0 & 100.0 & 100.0 & 0.0 & 0.0 \\
\hline
\end{tabular}
\caption{Wrong-entity action rate by ambiguity condition. Values are reported as percentages. Ent. Ret. denotes entity retrieval, Conf. denotes confidence-gated binding, and Ent. CMTF denotes entity-aware CMTF with provenance.}
\label{tab:ambiguity_results}
\end{table}

\subsection{Model-Level Trends}

Wrong-entity failures appear across all evaluated model families, although rates vary by backend. This indicates that entity binding is not merely a weakness of a single model. Importantly, every model has zero wrong-tool errors in the aggregate evaluation, reinforcing that the observed failures are specifically entity-binding failures rather than tool-selection failures.

We do not interpret model-level differences as stable rankings because model behavior may change with prompting, runtime configuration, and provider updates. The more important finding is qualitative: across model backends, action-oriented methods can produce wrong-entity actions even when wrong-tool errors are absent, while entity-aware execution policies prevent wrong-entity execution in this diagnostic setting.

\subsection{Representative Wrong-Entity Errors}

Representative logged failures show several recurring patterns. In temporal calendar tasks, models often call the correct rescheduling tool but bind it to an internal launch-sync event rather than the intended launch event. In true-ambiguity document deletion tasks, models call the correct deletion tool but choose one plausible old launch-plan document even though multiple old launch-plan documents are available. In name-collision tasks, models call the correct email-sending tool but select the wrong Alex. These cases illustrate why tool correctness alone is insufficient: the outward API call may look valid, while the real-world target is wrong.

\subsection{Summary of Findings}

The experimental results support three main findings. First, wrong-entity failures occur even when wrong-tool errors are eliminated. Second, retrieval and tool filtering alone do not reliably prevent wrong-entity actions. Third, entity-aware gating and provenance-based checks eliminate wrong-entity execution in this diagnostic setting, but introduce conservative deferral on unresolved ambiguous tasks. These findings motivate treating entity binding as a first-class reliability and safety layer in tool-augmented LLM agents.

\section{Discussion}
\label{sec:discussion}

The results show that entity binding failures are a distinct and practically important failure mode in tool-augmented LLM agents. In our diagnostic setting, all methods achieve a wrong-tool rate of 0.0\%, yet action-oriented methods still produce wrong-entity actions in roughly one quarter of runs. This separation is central: an agent can select the correct API, produce syntactically valid arguments, and appear to follow the user request while still acting on the wrong real-world target. In enterprise workflows, the harm is often attached not only to what action is taken, but to who or what the action is taken on.

\subsection{Tool Correctness Is Not Sufficient}

Most tool-use evaluations focus on whether the agent selects the correct tool, produces valid arguments, or completes the task. Our results show that these criteria are incomplete. A valid call to \texttt{send\_email} is not safe if the recipient is the wrong Alex. A valid call to \texttt{delete\_document} is not safe if the agent deletes the wrong launch plan. A valid call to \texttt{reschedule\_event} is not safe if the agent moves the wrong calendar event.

This distinction changes how tool-agent reliability should be measured. Tool correctness should be decomposed into at least two components: choosing the right operation and binding that operation to the right external entity. Evaluations that measure only tool selection or API validity can miss failures where the outward tool call is correct but the real-world target is wrong.

\subsection{Retrieval and Filtering Are Not Enough}

The results also show that exposing the agent to more relevant context does not automatically solve entity binding. Entity retrieval performs similarly to direct prompting in aggregate, producing a wrong-entity rate of 26.0\%. CMTF-only and semantic filtering improve this only slightly, with wrong-entity rates of 25.7\% and 24.0\%, respectively. These methods help organize the tool or context space, but they do not require the agent to verify that a specific entity binding is sufficiently supported before acting.

This suggests that entity binding is not merely a retrieval problem. Retrieval can surface candidate entities, but it does not decide whether the request uniquely identifies one of them. Similarly, tool filtering can remove irrelevant tools, but it does not resolve ambiguity among plausible entities. In ambiguous or near-ambiguous workflows, the agent needs an explicit execution policy for deciding when an entity is grounded strongly enough to permit action.

\subsection{Entity-Aware Gating Trades Completion for Safer Execution}

Entity-aware methods change the failure mode. In this diagnostic setting, confidence-gated binding and entity-aware CMTF with provenance eliminate wrong-entity actions and reduce risk-weighted wrong-entity exposure to 0.0. They do not achieve this by completing every task. Instead, they defer execution when the entity binding is unresolved.

This behavior is especially important for high-risk and critical actions. In true-ambiguity cases, no unique entity can be recovered without additional user input. Action-oriented baselines nevertheless execute the requested operation, while entity-aware methods treat ambiguity as a blocking condition. For actions such as deleting documents, cancelling events, modifying customer records, or sending externally visible communication, clarification is often preferable to silent execution on the wrong entity.

The tradeoff is lower direct task completion. Action-oriented methods achieve higher task success because they are evaluated under direct execution, but some of that completion comes from guessing. Entity-aware methods achieve lower task success because they refuse to act when the binding is insufficiently grounded. This should be interpreted as a safety--completion tradeoff rather than a single accuracy ranking. The relevant objective is calibrated execution: act when the binding is sufficiently grounded, and clarify when it is not.

A notable result is that over-clarification is 0.0\% in our evaluation. The entity-aware methods do not ask for clarification on unambiguous tasks. Their conservatism appears primarily on ambiguous or insufficiently grounded tasks, suggesting that entity-aware gating can reduce unsafe execution without necessarily adding unnecessary friction to clear requests.

\subsection{Where Entity Binding Fails}

The strongest wrong-entity effects appear in temporal and true-ambiguity tasks. Temporal tasks are difficult because the correct entity depends on contextual cues such as time, meeting purpose, participant set, recency, or event instance. True-ambiguity tasks are different: the request itself lacks enough information to identify a unique target. In those cases, any concrete action is unsafe unless the user clarifies.

Name collisions and cross-system references also produce wrong-entity failures. These cases are common in enterprise environments, where multiple people can share a first name, multiple documents can have similar titles, and the same project may appear in email, calendar, document, ticketing, and customer-record systems. The results suggest that entity binding should be treated as a cross-system grounding problem rather than a local string-matching problem.

\subsection{Implications for Agent Design}

The findings point to several design implications for tool-augmented agents. First, entity resolution should be represented explicitly in the agent control loop. Before executing a tool call, the agent should identify candidate entities, determine the intended binding, and assess whether the binding is sufficiently grounded.

Second, clarification should be treated as a valid safety-preserving outcome, not merely as failed task completion. If a user asks to ``delete the old launch plan'' and several old launch plans exist, the safe behavior is to ask which document they mean. Benchmarks and production systems should therefore distinguish between failed execution and safe deferral.

Third, provenance matters. High-impact actions should be linked to evidence. A binding decision should be traceable to the user request, retrieved candidates, entity attributes, and the reason a particular entity was selected or rejected. This can support auditing, debugging, and user trust.

Finally, tool filtering and entity binding should be combined rather than treated as substitutes. Tool filtering reduces the action space, while entity binding validates the target of action. An agent needs both. A minimal visible tool menu does not guarantee safe execution if the remaining tool is bound to the wrong entity.

Overall, the results motivate an explicit entity-aware layer between natural-language intent and external action. The central safety question is not only whether the agent called the right tool, but whether it bound that tool call to the right real-world entity and recognized when that binding was uncertain.

\section{Limitations and Threats to Validity}
\label{sec:limitations}

This study is designed as a controlled diagnostic evaluation rather than a comprehensive benchmark of all tool-augmented agent behavior. This design lets us isolate entity binding failures from tool-selection failures, but it also limits ecological validity. Real enterprise systems contain noisier, evolving, and partially observable entity stores, including stale records, incomplete metadata, permissions constraints, and cross-system inconsistencies. Because the environments and entity stores in this study are controlled, the reported rates should be interpreted as diagnostic failure rates under the constructed conditions rather than estimates of deployment-wide prevalence.

The task suite is also limited in size and scope. We evaluate 60 tasks across email, calendar, document, customer-record, and issue-tracking workflows. These domains are representative of common enterprise agent actions, but they do not cover every high-impact setting. Workflows involving payments, infrastructure operations, medical records, legal documents, or security administration may introduce different ambiguity patterns and risk profiles. Similarly, our ambiguity categories cover common cases such as name collisions, document-version ambiguity, temporal references, account collisions, near-duplicate records, cross-system references, and true ambiguity, but they do not exhaust all possible reference-resolution challenges.

The experiments focus on single-step tool execution decisions. This is appropriate for measuring whether an agent binds a requested action to the correct entity before execution. However, many deployed agents perform multi-step plans. In such settings, an early binding error may propagate through retrieval, reasoning, summarization, and subsequent tool calls. Multi-step agents may also introduce additional opportunities for verification, correction, or recovery. Our evaluation therefore measures first-step entity binding failures rather than full workflow failure rates.

The results may depend on prompting, output format, decoding behavior, and model version. We use a structured JSON response format and a fixed prompting regime to make methods comparable across models. In particular, the action-oriented baselines are evaluated under a direct-execution policy, while entity-aware methods are allowed to clarify or defer. The comparison should therefore be read as measuring the safety--completion tradeoff between action-first and entity-aware execution policies, not as proving that the baselines could not be made safer with different prompts, explicit clarification rules, or additional execution gates. For the same reason, model-level results should not be read as permanent rankings of specific model backends.

The entity-aware methods are also diagnostic implementations rather than final production designs. The confidence-gated methods use structured prompting and candidate comparison rather than a separately calibrated entity-linking or uncertainty-estimation model. As a result, the confidence and margin behavior should be viewed as an abstract execution policy instantiated for controlled evaluation, not as a fully calibrated uncertainty estimator. Future systems may require learned scoring models, domain-specific thresholds, richer provenance representations, and stronger calibration under distribution shift.

Finally, our metrics reflect a safety-oriented interpretation of agent behavior. We count clarification as safe success when the task is truly ambiguous, because asking the user is preferable to guessing in high-risk actions. Other applications may assign higher cost to clarification, latency, or interruption. We evaluate whether the agent chooses to clarify, but we do not conduct a user study measuring whether the clarification is understandable, minimally burdensome, or sufficient for resolving the task. Likewise, ambiguity annotations and risk weights are useful for diagnostic comparison, but different annotators or deployment contexts may disagree about when a reference is sufficiently clear to permit execution or how severe a wrong-entity action should be.

Despite these limitations, the study identifies a concrete threat to validity in existing tool-use evaluations: measuring only whether an agent selects the right tool can miss failures where the agent applies that tool to the wrong real-world entity. This motivates treating entity binding as a first-class evaluation and execution requirement for dependable tool-augmented LLM agents.

\section{Conclusion}
\label{sec:conclusion}

Tool-augmented LLM agents are commonly evaluated by whether they choose the correct tool, produce valid arguments, and complete the requested task. This paper shows that these criteria are insufficient. An agent can select the right tool and still fail by binding that tool call to the wrong real-world entity. We define these errors as \emph{entity binding failures} and argue that they are a distinct reliability and safety problem for agents that operate over people, documents, calendar events, customer records, tickets, and other external entities.

We formalized the separation between tool correctness and entity correctness, introduced a taxonomy of common entity binding failures, and evaluated several tool-use methods in a controlled diagnostic testbed. Across 60 diagnostic tasks, five model backends, and six tool-use methods, wrong-tool error was 0.0\%, yet action-oriented methods still produced wrong-entity actions in 24.0--26.0\% of runs. This demonstrates that correct tool selection does not imply safe execution. Retrieval, semantic filtering, and CMTF-style tool filtering can organize the tool or context space, but they do not by themselves ensure that the selected action is grounded to the correct entity.

Entity-aware execution policies changed this behavior. In this diagnostic setting, confidence-gated binding and entity-aware CMTF with provenance eliminated wrong-entity actions by blocking execution when the entity binding was insufficiently grounded. This introduces a safety--completion tradeoff: conservative methods complete fewer tasks directly, but avoid unsafe execution under ambiguity. For high-impact workflows, this tradeoff is often desirable. Asking for clarification is not merely a fallback behavior; it is a necessary safety mechanism when the intended entity cannot be uniquely determined.

The broader implication is that dependable tool use requires an entity-aware layer between natural-language intent and external action. Future agent systems should explicitly retrieve candidate entities, evaluate binding confidence, track provenance, and block execution when the target entity is uncertain. Future benchmarks should likewise measure not only whether an agent called the right tool, but whether it acted on the right entity and recognized when that binding was ambiguous.

Entity binding should therefore be treated as a first-class requirement for reliable tool-augmented LLM agents. As agents become more deeply connected to enterprise systems, the central safety question is not only what action the agent takes, but who or what the action is taken on.

\section*{Acknowledgment}
The authors thank colleagues for helpful feedback. This work was conducted in the authors' personal capacity. The views expressed in this paper are solely those of the authors and do not necessarily reflect the views of their employers. This work did not receive external funding. The authors declare no conflicts of interest.

\section*{Artifact Availability}

The benchmark, tool registry, filtering implementations, evaluation scripts, prompts, and analysis utilities used in this study are publicly available at: \url{https://github.com/R-Suresh/EntityBindingFailures}. The repository contains task definitions, evaluation harnesses, and scripts required to reproduce the reported results.

\bibliographystyle{IEEEtran}
\bibliography{references}

\end{document}